\documentclass[runningheads]{llncs}
\usepackage{amsmath,amsfonts}
\usepackage{mathtools}
\usepackage{comment}
\usepackage{graphicx}
\usepackage{cite}
\usepackage{tikz,pgfplots}
\usepackage[export]{adjustbox}
\usepgfplotslibrary{groupplots}
\pgfplotsset{compat=newest}
\usetikzlibrary{arrows,backgrounds,positioning}
\usetikzlibrary{decorations.pathmorphing}
\usepgfplotslibrary{colorbrewer}
\usepackage{url}
\usepackage[mathscr]{eucal}
\usepackage[noend]{algpseudocode}
\usepackage[ruled]{algorithm}

\newcommand{\bs}{\mathbf}
\newcommand{\isdef}{\mathrel{\mathrel{\mathop:}=}}

\def\letters{a,b,c,d,e,f,g,h,i,j,k,l,m,n,o,p,q,r,s,t,u,v,w,x,y,z}
\def\Letters{A,B,C,D,E,F,G,H,I,J,K,L,M,N,O,P,Q,R,S,T,U,V,W,X,Y,Z}
\makeatletter
\@for \@l:=\Letters \do{%
  \expandafter\edef\csname\@l bb\endcsname{\noexpand\ensuremath{%
  \noexpand\mathbb{\@l}}}%
  \expandafter\edef\csname\@l bf\endcsname{{\noexpand\bf \@l}}%
  \expandafter\edef\csname\@l cal\endcsname{\noexpand\ensuremath{%
  \noexpand\mathcal{\@l}}}%
  \expandafter\edef\csname\@l eu\endcsname{\noexpand\ensuremath{%
  \noexpand\EuScript{\@l}}}%
  \expandafter\edef\csname\@l frak\endcsname{\noexpand\ensuremath{%
  \noexpand\mathfrak{\@l}}}%
  \expandafter\edef\csname\@l rm\endcsname{{\noexpand\rm \@l}}%
  \expandafter\edef\csname\@l scr\endcsname{\noexpand\ensuremath{%
  \noexpand\mathscr{\@l}}}%
}
\@for \@l:=\letters \do{%
  \expandafter\edef\csname\@l bf\endcsname{{\noexpand\bf \@l}}%
  \expandafter\edef\csname\@l frak\endcsname{\noexpand\ensuremath{%
  \noexpand\mathfrak{\@l}}}%
  \expandafter\edef\csname\@l scr\endcsname{\noexpand\ensuremath{%
  \noexpand\mathscr{\@l}}}%
}
\begin{document}
\title{Observation-specific explanations through scattered
data approximation}
%
%
\author{Valentina Ghidini \and Michael Multerer \and
Jacopo Quizi \and
Rohan Sen}
\authorrunning{V. Ghidini et al.}
%
\institute{Istituto Eulero, Universit\`a della Svizzera Italiana\\
Lugano, Svizzera\\
\email{\{name.surname\}@usi.ch}}
\maketitle              
%
\begin{abstract}
This work introduces the definition of observation-specific 
explanations to assign a score to each data point proportional to its
importance in the definition of the prediction process. 
Such explanations involve
the identification of the most influential observations for the
black-box model of interest. The proposed method involves estimating 
these explanations by constructing a surrogate model through 
scattered data approximation utilizing the orthogonal matching pursuit 
algorithm. The proposed approach is validated on both simulated and real-world datasets.             

\keywords{Observation-specific explanations \and functional reconstruction 
\and surrogate model \and reproducing kernel Hilbert space.}
\end{abstract}

\section{Introduction}\label{sec:introduction}
In the rapidly evolving landscape of machine learning, the need for 
understanding and interpreting complex models has become increasingly important. 
As models get more sophisticated and are applied to critical decision-making 
processes across various domains, the demand for transparency and 
interpretability has never been higher. This necessity has led to the development of
explainable artificial intelligence (XAI), which aims to shed light on the inner
workings of black-box algorithms.

Most of the conventional XAI techniques found in the current literature, such as 
the ones in \cite{LundLee17,Guidotti2018,
Ribeiro2016,Petsiuk2018,Molnar2019,borgonovo2023,ghidini2023_xai}, 
can be broadly classified as measures of variable importance. The common rationale of all these 
techniques is to assign a score to the different input features according to their respective  importance in the inner mechanisms of the black-box model to 
explain. In fact, the higher the score, the more significant the variable. This approach
is very general and versatile, as these methods can be applied to different data
formats, including images and texts, and are compatible with diverse types of
black-box models.

In this work, however, we propose to shift the perspective from explanatory
variables to individual data points. Specifically, our objective is to
understand the importance of each data instance in estimating the 
decision-making process. We anticipate that some data points may have a greater
impact than others on the definition of the predictions. This change of focus
offers an opportunity to better understand the internal operations of a model.
By detecting the most important observations, we can analyze what are the influential data
instances driving the estimation of the black box and understand whether the
model accurately captures the ``typical'' data points or is ill-fit to represent
outliers or unexpected instances. In essence, this approach enables us to
identify the data points that have the greatest influence on the decision-making
process and draw pertinent conclusions accordingly.

The idea proposed in this paper is closely related to influential observations
in statistical models~\cite{belsley2005,chatterjee1986}. Influential
observations are conceptually different from outliers: in the literature, they
are broadly defined as data points whose impact on the model estimation
significantly outweighs that of others. In linear models, influential data
points are well-defined and diagnosed through metrics like leverage and
Cook's distance~\cite{cook1979}. However, identifying influential observations
in black boxes is still an open problem.

We begin by providing a proper definition of
observation-specific explanations. These explanations are designed to measure 
the influence of each data point on the decision-making process of the black 
box. Then, we propose to estimate such explanations using the scattered data
approximation~\cite{Wen05},
effectively defining an interpretable surrogate model for the black box.
If the surrogate lives in a reproducing kernel Hilbert space, it can
be represented with respect to an orthonormal basis. This allows us to use the resulting
coefficients in the expansion to define normalized observation-specific explanations.
The surrogate model is estimated using the orthogonal matching pursuit
algorithm~\cite{PS2011,MS09,multerer2024fast} that 
selects an optimal subset of data points to represent the black-box model of 
interest. The resulting predictions of the black box at the selected data
instances are perfectly reproduced by the surrogate, and the approximation error at all data points is rigorously bounded.
This even enables
an observation-wise diagnostic of the fidelity of the surrogate model~\cite{Molnar2019}. Then, the approach is validated on synthetic data and on a data set concerning anatomical measurements of animals.

The remainder of the paper is organized as follows:
Section~\ref{sec:methodology} introduces the definition of observation-specific
explanations and proposes an estimation procedure leveraging
orthogonal matching pursuit in reproducing kernel Hilbert spaces.
Section~\ref{sec:application} presents the application of the proposed method 
on both simulated scenarios and real datasets. Finally, 
Section~\ref{sec:discussion} summarizes the findings and discusses future
research directions.

\section{Methodology}\label{sec:methodology}
We adopt a standard machine-learning notation. Consider a sample set of $n$ observations, each containing $p$ features, denoted as $X = \{\bs x_1, \ldots, \bs x_n\}\subset\mathcal{X} \subseteq \Rbb^p$, where $\Xcal$ is the sample space.
We denote by $f$ the black-box model 
of interest which outputs the predictions $f(\bs x_1), \ldots,f( \bs x_n)$ at the sample points.
Given this setting, we begin by defining observation-specific explanations
that offer instance-wise significance measures proportional to the importance of each data point in the definition of the prediction process. Following this
idea, we introduce a framework for computing such explanations using a
surrogate model based on scattered data approximation techniques.

\subsection{Observation-specific explanations}
\begin{definition}\label{def:obs_explanations}
     Given a model $f$, we define \emph{observation-specific explanations} as 
    \[
    \gamma = \{ \gamma_1, \dots, \gamma_n \},
    \]
    where $\gamma_i \in[0,\infty), \, 1 \leq i \leq n,$ is a positive index proportional to the importance of 
    the $i^{th}$ observation for the model $f$.
\end{definition}

To the best of our knowledge, this definition introduces a
novel class of explanations in the literature. It proposes a family of
importance metrics, indicating how much each observation influences
the predictions made by the black-box model under examination.  The viewpoint is 
then shifted with respect to standard explainability techniques: indeed, 
traditionally, importance is assessed for each explanatory variable regarding 
the prediction process, typically by averaging across dataset observations. 
However, our proposal reverses the approach: the aim is to understand the 
impact of each data point on the process of generating the predictions. As mentioned above, this concept draws inspiration from influential
observations in statistical models, which are intuitively defined as instances
whose removal or addition significantly impacts the model 
estimation~\cite{belsley2005}.

Definition~\ref{def:obs_explanations} is kept general since it can encompass any specific algorithm that computes the observation-specific explanations.  
To address the latter, we propose a procedure based on the
definition of a surrogate model using scattered data approximation in the next subsection. 
However, future research directions could explore the development of
alternative algorithms to provide observation-specific explanations. 

\subsection{Surrogate models using scattered data approximation}
To estimate observation-specific explanations for a particular black-box function~$f$, we define a surrogate model~$f^*$ such that $f^*(\mathbf{x}) \approx f(\mathbf{x})$ 
for~$\bs x \in \mathcal{X}$. 
Recall that a surrogate model is a function designed to closely mimic the 
behavior of the black-box model under consideration
\cite{Ribeiro2016,Molnar2019}.

Suppose that the surrogate model $f^\ast$ is contained in a reproducing kernel Hilbert space (RKHS)~\(\Hscr\)
that contains real-valued functions on the sample space~$\Xcal$ with reproducing kernel \(\Kscr\colon\Xcal\times\Xcal\to\Rbb\), see~\cite{RKHS_book}. \textcolor{black}{This assumption implies that the surrogate is defined as a linear model as in Equation~\eqref{eq:kernelApprox} in the RKHS~$\Hscr_X$}.
Using the canonical feature map $\bs x \mapsto \Kscr(\bs x, \cdot) \in \Hscr$, the sample set~\(\{{\bf x}_1,\ldots,{\bf x}_n\}\) can be embedded into~$\Hscr$.
We \textcolor{black}{posit} that the black-box model $f$ can be well approximated by the surrogate model $f^\ast$, that is
\begin{equation}\label{eq:kernelApprox}
 f({\bf x})\approx f^\ast(\bs x) = 
 \sum_{i=1}^n c_i\,\Kscr({\bf x}_i, \bs x)\quad \text{for all }
 {\bf x} \in \mathcal{X}.
\end{equation}
\textcolor{black}{Here, the $c_i$'s are the coefficients corresponding to the kernel function $\Kscr(\bs x_i,\cdot)$ parametrizing the linear representation of the surrogate model.}
From \eqref{eq:kernelApprox}, it is possible to derive a class of surrogate 
models that exactly match the values of the predictions 
$f({\bf x}_{i_1}), \dots, f({\bf x}_{i_{n^\ast}})$ at a selected subsample~$\{{\bf x}_{i_1}, \ldots, {\bf x}_{i_{n^\ast}}\} 
\subset \{{\bf x}_1,\ldots,{\bf x}_n\}$.
To compute this subsample, we adaptively compute an orthonormal basis
\(\{\Nscr_1,\ldots,\Nscr_{n^\ast}\}\) of \(\Hscr_X\)
by means of \emph{orthogonal matching pursuit}~\cite{PS2011} (OMP). 
This approach amounts to representing the surrogate model with respect to this new basis:
\begin{equation}\label{eq:interpolant_orthonormal}
    f^\ast (\bs x)= \sum_{j=1}^{n^\ast} a_j^{\ast}\,\Nscr_j({\bf x}).
\end{equation}
\textcolor{black}{Now, the $a^\ast_j, j = 1, \dots, n^\ast$, are the coefficients with respect to the new (orthonormal) basis functions. However, we cannot use these coefficients directly for observation-specific explanations since any particular basis function $\Nscr_j(\bs x)$ is not uniquely linked to a single observation.}
Nevertheless, the orthonormal basis functions~\(\Nscr_1,\ldots,\Nscr_{n^\ast}\) can be computed
by a Gram-Schmidt procedure with respect to a certain reordering
of the basis of kernel translates \(\{\Kscr({\bf x}_1,\cdot),\ldots,
\Kscr({\bf x}_n,\cdot)\}\).
Therefore, we obtain
\begin{equation}\label{eq:Newton_rep}
    \Nscr_j({\bf x})=\sum_{k=1}^j d_{k,j}\Kscr({\bf x}_{i_k},{\bf x})
\end{equation}
for certain coefficients \(d_{k,j}\). Thus, \textcolor{black}{Equation~\eqref{eq:Newton_rep} can be used to represent the surrogate in terms of the kernel basis functions, which can subsequently be used for observation-specific explanations.} Notice that \textcolor{black}{the indices of the selected observations} can be obtained in an iterative manner. Indeed, having computed \(m<n^*\)
of these basis functions, the next index \(i_{m+1}\) is chosen such that
the error
\begin{equation}
    \max_{i=1,\ldots,n}\bigg|f({\bf x}_i)-
\sum_{j=1}^{m} a_j^{\ast}\,\Nscr_j({\bf x}_i)\bigg|
\end{equation}
is minimized. 

A surrogate model such as the above is particularly advantageous for two main reasons. First, using \eqref{eq:Newton_rep}, we can express $f^\ast$ in terms of $\Kscr(\bs x_1,\cdot), \ldots, \Kscr(\bs x_n,\cdot)$ to define quantities that can be interpreted as observation-specific explanations. In particular, we can use the respective coefficients of the kernel basis functions to assign an appropriate weight to each observation proportional to its significance. It is worth noting that each kernel basis function corresponds to a particular sample point. Thus, the OMP effectively selects a subsample of $n^\ast$ data points for the reconstruction process, which are then targeted as the most influential observations. Second, it offers analytical bounds on the reconstruction of 
$f({\bf x}_i)$ for \(i=1,\ldots,n\). This implies that within a standard
mathematical framework, there are theoretical guarantees that~$f^*$ accurately 
replicates the behavior of the black-box model $f$ on \(X\). Furthermore, if we assume that $f \in \Hscr$, then we can also have the guarantees of a high fidelity on the entire sample space \(\Xcal\).
It is worth noting that such an approach allows us to obtain observation-specific errors, defined as
\begin{equation}\label{eq:errors_obs}
    \text{err}_i \isdef \vert f(\bs x_i) - f^\ast(\bs x_i) \vert, \qquad \text{for } i =1, \dots, n.
\end{equation}
In general, it can also be proven that using a tolerance $\varepsilon^2$ in the OMP algorithm proposed in~\cite{multerer2024fast}, there exists the uniform error bound
\begin{equation}\label{eq:worst_case_error}
    \vert f(\bs x_i) - f^\ast(\bs x_i) \vert \leq \varepsilon \, \|f\|_{\Hscr}, \quad \text{for } i =1, \dots, n.
\end{equation}
Notice that the tolerance can be interpreted as the extent to which we want to reconstruct the black-box model using the surrogate model and gives us uniform error bounds as in~\eqref{eq:worst_case_error}. \textcolor{black}{In general, given a tolerance $\varepsilon$, the number of selected data points $n^\ast$ depends on the overall distribution of the observations in the support as well the type of kernel used for the surrogate model. As a result, the choice of the kernel for the reconstruction process is of paramount importance. Future research work should assess the sensitivity and study the nature of such a choice.}
Analyzing such errors \textcolor{black}{(and their normalized versions)}, we can discern for which observations the prediction process is accurately represented by the surrogate, i.e., where the error is close or equal to zero and where this is not the case. Thus, we can also sketch a tentative analysis of the surrogate model to see whether there are some systematic areas of the point cloud bound for errors.

\subsection{Estimation of the observation-specific explanations}
A representation as in \eqref{eq:interpolant_orthonormal} lends itself to offering observation-specific explanations. Note that the expression of the surrogate model $f^\ast$ in terms of an orthonormal basis implies that each coefficient $a_j^\ast$ is simply the projection of the black-box model $f$ onto the corresponding basis function pertaining to a particular observation from the set of samples. We wish to use this characterization of the coefficients to identify the most influential observations.
Using \eqref{eq:Newton_rep} in \eqref{eq:interpolant_orthonormal}, we have that 
\begin{equation}
    f^\ast(\bs x) = \sum_{j=1}^{n^\ast} a_j^\ast \sum_{k=1}^j d_{k,j} \Kscr(\bs x_{i_k}, \bs x) 
    = \sum_{k=1}^{n^\ast} c_{i_k} \Kscr(\bs x_{i_k}, \bs x),
\end{equation}
where \[
c_{i_k} = \sum_{j=k}^{n^\ast} d_{k,j} a_j^\ast.
\]
\textcolor{black}{Each coefficient $c_{i_k}$ is then uniquely associated to a selected observation.} Denoting the set $S\isdef \{i_1,\ldots, i_{n^\ast}\}$, we set $c_i=0$ if $i\notin S$ for all $1\leq i \leq n$. Finally, we define the observation-specific explanations as
\begin{equation}\label{eq:obs_explanations}
\gamma_i = 
\dfrac{|c_i|}{\underset{1\leq i \leq n}{\max}|c_i|}.
\end{equation}
Thus, the data points not deemed as important are associated with a null explanation, effectively annihilating their impact on the estimation of the surrogate model, and consequently on the prediction process provided by the black box. The resulting values of $\gamma_i$ serve as normalized measures of the importance of each observation: a higher $\gamma_i$ indicates greater importance of the $i^{th}$ data point for the prediction process of the black box, replicated by the surrogate model.

\section{Application}\label{sec:application}

In this section, we showcase the application of the proposed method. In particular, Subsection~\ref{subsec:simulation} displays results in simulated scenarios, using well-known data-generating processes. Then, Subsection~\ref{subsec:realdata} presents an application on a real dataset. 
 
\subsection{Simulated studies}\label{subsec:simulation}
In this subsection, we will explain two known data-generating processes that are suitably defined, rather than dealing with an actual black-box model. This framework facilitates assessing the suitability of the surrogate model and understanding the features of the resulting observation-specific explanations in a well-understood context.

\begin{figure}[htbp!]
\begin{center}
\scalebox{0.9}{
\begin{tikzpicture}
\begin{axis}[
grid = both,
ticklabel style = {font=\tiny, ultra thick},
colorbar horizontal,
colorbar style={
at={(0.5,0.85)},anchor=north },
   colorbar style={
         point meta min=0,  
         point meta max =4.2e-3,
         scaled x ticks = false,
        xtick = {0,  4.2e-3},
            width = 20mm,
            title = Error,
            title style={xshift=-0.6cm,yshift=-0.25cm},
            },
         colorbar/width=2.5mm,
y label style={ anchor=center },
xlabel=$X_1$,
y label style={rotate=75},
label style={font=\tiny},
zticklabel pos=left,
ylabel=$X_2$,
zlabel = {$f(X_1\text{,}X_2)$},
grid style={ultra thin, gray!15},
x label style= {rotate=90},
  every axis x label/.append style={sloped, at={(rel axis cs: 0.5, -0.250, 0)}, below},
  every axis y label/.append style={sloped, at={(rel axis cs: 1.250, 0.5, 0)}, below}, 
colormap/YlOrRd,
point meta min = 0,
view={45}{30},
height=6.85cm,width=6.85cm,
mark size=1pt
]
\addplot3+[only marks, mark size=0.7pt, mark = *,scatter,scatter/use mapped color={draw=mapped color, fill=mapped color},
  scatter,
  scatter src=explicit,
  mark options = fill] table [x = X1,y = X2, z= Y_pred, meta expr= {(abs(1e15*(\thisrow{Abs_err})))}]{test_generate_data_squared.txt};
 \end{axis}
\end{tikzpicture}}
\scalebox{0.9}{
\begin{tikzpicture}
\begin{axis}[
grid = both,
legend style={at={(0.5,.85)},anchor=north},
 legend style={draw=none},
ticklabel style = {font=\tiny, ultra thick},
y label style={ anchor=center },
xlabel=$X_1$,
y label style={rotate=75},
label style={font=\tiny},
zticklabel pos=left,
ylabel=$X_2$,
zlabel = {$f(X_1\text{,}X_2)$},
grid style={ultra thin, gray!15},
x label style= {rotate=90},
  every axis x label/.append style={sloped, at={(rel axis cs: 0.5, -0.250, 0)}, below},
  every axis y label/.append style={sloped, at={(rel axis cs: 1.250, 0.5, 0)}, below}, 
view={45}{30},
height=6.85cm,width=6.85cm,
legend image post style={scale=3},
mark size=1.5pt,
]
\addplot3+[only marks, mark size=0.7pt, mark = *,
  color = lightgray!30,
  mark options={fill=lightgray!30}
  ]
  table [ x = X1,y = X2, z= Y_pred ] {test_generate_data_squared.txt};
  \addplot3+[only marks, colormap/YlOrRd,,mark size=0.7pt, mark = *,scatter,scatter/use mapped
color={
    draw=mapped color,
    fill=mapped color,
  },
  scatter,
  point meta min = 0,
      visualization depends on=\thisrowno{3}\as\wtwo,
    scatter/@pre marker code/.append style={
        /tikz/mark size=0.5*ln(exp(1)+50*abs(\wtwo))},
  scatter src=explicit,
  mark options = fill]
    table[x index = 0, y index = 1, z index = 2,meta expr= {(abs(\thisrowno{3}))}] {test_generate_data_squared_expl.txt};
    \legend{data, selected}
\end{axis}
\end{tikzpicture}}
\caption{First simulated scenario: the data generating process is a quadratic function. In the left plot, the colors correspond to the observation-specific relative error in the surrogate model reconstruction: darker shades represent higher errors. In the right plot, selected data points are colored and sized according to the magnitude of their explanations: Darker and larger points indicate higher values.}
\label{fig:sim_quadratic}
\end{center}
\end{figure}
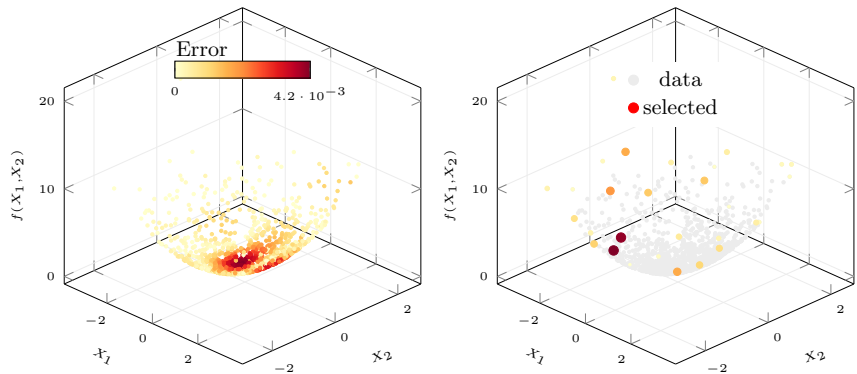

The first example concerns a simple, bidimensional, quadratic function. We simulate $n=1000$ data points according to
\begin{equation}\label{eq:sim_dgp_quadratic}
    f( X_1, X_2)= X_1^2 + X_2^2 + 1,
\end{equation}
where $X_1$ and $X_2$ are random, independent, standard Gaussian explanatory variables.
We construct the surrogate model in~\eqref{eq:interpolant_orthonormal} by employing a Gaussian kernel, see \cite{SVM_book}, with a length scale parameter equal to $\sqrt{7}$. We discuss an alternative method for determining this crucial hyperparameter using the real dataset in Subsection~\ref{subsec:realdata}. The tolerance is set to $\varepsilon = 10^{-3}$, and the procedure selects $n^* = 36$ data points for the definition of the surrogate model. In particular, the reconstruction error in~\eqref{eq:worst_case_error} is of the order of $10^{-4}$, indicating that we are achieving a nearly perfect reconstruction of the real function using only $3.6\%$ of the data size: this means that the surrogate model has a provable high fidelity in replicating the generating model of interest.
Figure~\ref{fig:sim_quadratic} reports the results on the bidimensional, quadratic simulated dataset: the left plot displays the data points with colors proportional to the observation-specific errors in Equation~\eqref{eq:errors_obs} --- darker hues indicate higher errors and consequently a lower fidelity of the surrogate model. 
This visualization offers insight into the data points best represented by the surrogate model. The surrogate reproduces better the instances at the data cloud's boundaries where the points' density is lower. These boundary points are also where we anticipate having statistically more uncertainty in the functional reconstruction, but also perfect candidates for interpolation. 
In the right plot, the colored points represent those selected for defining the surrogate in Equation~\eqref{eq:interpolant_orthonormal}, with the intensity of red and the point size proportional to the observation-specific explanations in~\eqref{eq:obs_explanations}. As expected, such instances shape the actual point cloud, focusing on more representative areas of the support. 

\begin{figure}[htbp]
\begin{center}
\scalebox{0.9}{
\begin{tikzpicture}
\begin{axis}[
grid = both,
ticklabel style = {font=\tiny, ultra thick},
colorbar horizontal,
colorbar style={
at={(0.45,0.25)},anchor=north },
   colorbar style={
         point meta max =0.1,
         scaled x ticks = false,
        xtick = {0,  0.1},
            width = 20mm,
            title = Error,
            title style={xshift=-0.6cm,yshift=-0.25cm},
            },
         colorbar/width=2.5mm,
y label style={ anchor=center },
xlabel=$X_1$,
y label style={rotate=90},
label style={font=\tiny},
zticklabel pos=left,
ylabel=$X_2$,
zlabel = {$f(X_1\text{,}X_2)$},
grid style={ultra thin, gray!15},
x label style= {rotate=90},
  xlabel style={sloped like x axis},
    ylabel style={sloped like y axis},
  every axis x label/.append style={sloped, at={(rel axis cs: 0.5, 1.25, 0)}, below},
  every axis y label/.append style={sloped, at={(rel axis cs: -0.25, 0.5, 0)}, below}, 
colormap/YlOrRd,
view={225}{30},
height=6.85cm,width=6.85cm,
mark size=.7pt,
]
\addplot3+[only marks, mark size=0.7pt, mark = *,scatter,scatter/use mapped
color={
    draw=mapped color,
    fill=mapped color,
  },
  scatter,
  scatter src=explicit,
  mark options = fill] table [ x = X1,y = X2, z= Y_pred, meta expr= {sqrt(sqrt(abs((\thisrow{Abs_err}))))}] {test_generate_data_ackley.txt};
 \end{axis}
\end{tikzpicture}}
\scalebox{0.9}{
\begin{tikzpicture}
\begin{axis}[
grid = both,
legend style={at={(0.5,.25)},anchor=north},
 legend style={draw=none},
ticklabel style = {font=\tiny, ultra thick},
ticklabel style = {font=\tiny, ultra thick},
y label style={ anchor=center },
xlabel=$X_1$,
y label style={rotate=75},
label style={font=\tiny},
zticklabel pos=left,
ylabel=$X_2$,
zlabel = {$f(X_1\text{,}X_2)$},
legend style={font=\small},
grid style={ultra thin, gray!15},
x label style= {rotate=90},
  every axis x label/.append style={sloped, at={(rel axis cs: 0.5, -0.250, 0)}, below},
  every axis y label/.append style={sloped, at={(rel axis cs: 1.250, 0.5, 0)}, below}, 
view={45}{30},
height=6.85cm,width=6.85cm,
legend image post style={scale=3},
mark size=.7pt,
]
\addplot3+[only marks, mark size=0.7pt, mark = *,
  color = lightgray!30,
  mark options={fill=lightgray!30}
  ]
  table [ x = X1,y = X2, z= Y_pred ] {test_generate_data_ackley.txt};
  \addplot3+[only marks, colormap/YlOrRd,mark size=0.7pt, mark = *,scatter,scatter/use mapped
color={
    draw=mapped color,
    fill=mapped color,
  },
  scatter,
  point meta min = 0,
  scatter src=explicit,
        visualization depends on=\thisrowno{3}\as\wtwo,
    scatter/@pre marker code/.append style={
        /tikz/mark size=0.5*ln(exp(1)+50*abs(\wtwo))},
  mark options = fill]
    table [x index = 0, y index = 1, z index = 2,meta expr= {(abs(\thisrowno{3}))}]  {test_generate_data_ackley_expl.txt};
    \legend{data, selected}
\end{axis}
\end{tikzpicture}}
\caption{Second simulated scenario: the data generating process is the Ackley function. In the left plot, the colors correspond to the observation-specific relative error in the surrogate model reconstruction: darker shades represent higher errors. In the right plot, selected data points are colored and sized according to the magnitude of their explanations: Darker and larger points indicate higher values.}
\label{fig:ackley} 
\end{center}
\end{figure}
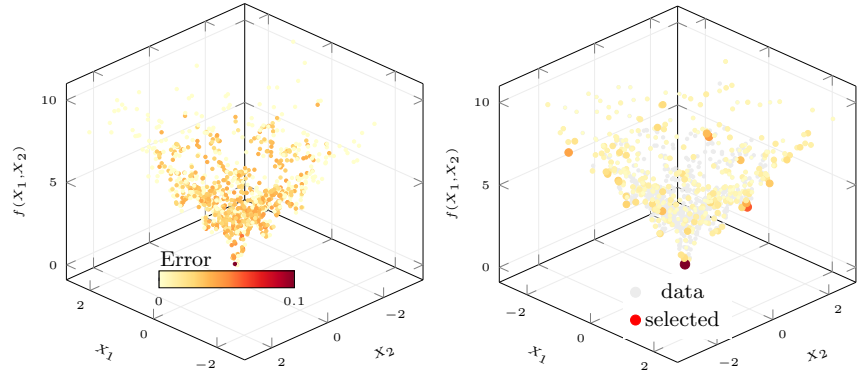

In the second simulated scenario, we employ the well-known Ackley function~\cite{ackley1987}, commonly used as a benchmark for optimization studies due to its highly oscillatory nature and the presence of multiple local optimal points. We opt for this function to explore how the surrogate model in~\eqref{eq:interpolant_orthonormal} behaves in the presence of a smooth yet intricate generating model, producing peculiarly shaped point clouds.
To estimate the surrogate model, we run the OMP algorithm choosing a tolerance of $\varepsilon = 10^{-3}$, which results in $n^* = 370$ selected points out of the $n=1000$ simulated ones. We simulate~$X_1, X_2$ from independent standard Gaussian distributions and employ the Matérn kernel \(\mathcal{K}_{3/2}\), see \cite{baroli2023samplet}. The resulting reproduction error is of the order of~$10^{-4}$.
Figure~\ref{fig:ackley} reports the results for this scenario.
 As before, the plot on the left illustrates the cloud of the simulated data points, with colors indicating the observation-specific relative error in~\eqref{eq:errors_obs}. Lighter points denote a better approximation (higher fidelity) by the surrogate model. Consequently, the boundaries of the point cloud reveal the lower fidelity of the surrogate, as these areas are statistically the most challenging to reproduce. However, even in these low-fidelity regions, we still select representative subsamples that can be used for effective reconstruction.
In the right plot, the points chosen by the OMP algorithm to construct the surrogate model are displayed. Their colors and sizes are proportional to the observation-specific explanation in~\eqref{eq:obs_explanations}, with darker shades and bigger points representing higher values. Notably, these selected data instances mostly correspond to the local optima of the Ackley function, heuristically selected for their informativeness.
Furthermore, the points with higher values of explanations are strategically positioned in less dense areas of the point cloud.

\subsection{Real-world application}\label{subsec:realdata}
In this section, we present an application of the proposed technique on a publicly available dataset\footnote{\url{kaggle.com/datasets/abrambeyer/openintro-possum}}, whose task is to predict the height of~$n=101$ possums given their anatomical features, including the skull size, foot length, and others, standardized. We consider a black-box model by training an AdaBoost regressor~\cite{esl}, achieving a mean square error of $3.64$ centimeters on the predicted length of the possum. This error is deemed acceptable since the mean length of the animals in this dataset is roughly $87$ centimeters, with a standard deviation of $13.23$ centimeters. We run the OMP algorithm, setting the tolerance $\varepsilon = 10^{-2}$ that gives us the number of selected data points to be equal to $n^* = 20$. Here, we choose the length scale parameter for the Exponential kernel, see \cite{baroli2023samplet}, with K-fold cross-validation~\cite{esl}, where the parameter is set to the minimizer of the cross-validation error, which is found to be equal to $9.32\cdot 10^{-4}$. The final, reconstruction error is of the order of $10^{-3}$. 

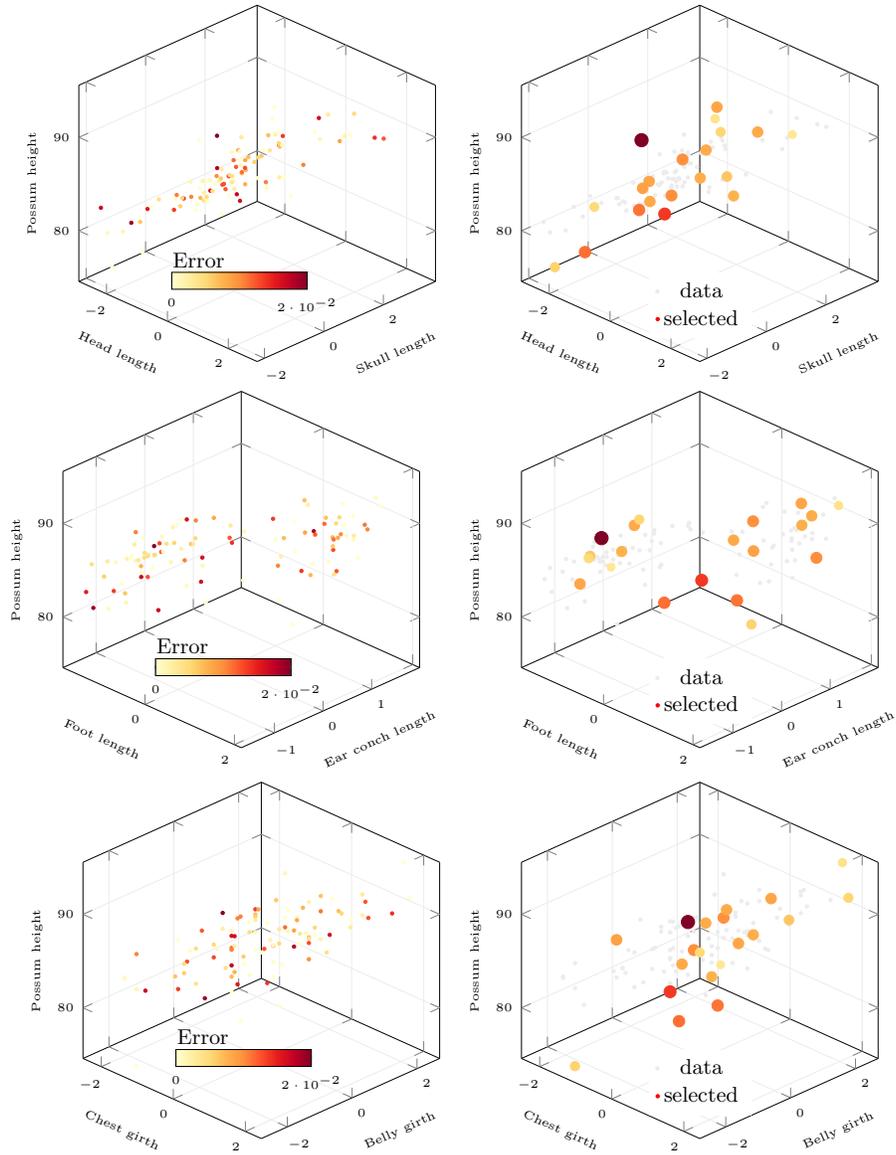
\begin{figure}[htbp!]
\begin{center}
\scalebox{0.9}{
\begin{tikzpicture}
\begin{axis}[
grid = both,
ticklabel style = {font=\tiny, ultra thick},
colorbar horizontal,
colorbar style={
at={(0.45,0.25)},anchor=north },
   colorbar style={
         point meta min=0,  
         point meta max = 0.02,
         scaled x ticks = false,
        xtick = {0,  0.02},
            width = 20mm,
            title = Error,
            title style={xshift=-0.6cm,yshift=-0.25cm},
            },
         colorbar/width=2.5mm,
y label style={ anchor=center },
xlabel=Head length,
y label style={rotate=75},
label style={font=\tiny},
zticklabel pos=left,
ylabel=Skull length,
zlabel = {Possum height},
grid style={ultra thin, gray!15},
x label style= {rotate=90},
  every axis x label/.append style={sloped, at={(rel axis cs: 0.5, -0.250, 0)}, below},
  every axis y label/.append style={sloped, at={(rel axis cs: 1.250, 0.5, 0)}, below}, 
colormap/YlOrRd,
point meta min = 0,
view={45}{30},
height=6.85cm,width=6.85cm,
mark size=.7pt,
]
\addplot3+[only marks, mark size=0.7pt, mark = *,scatter,scatter/use mapped
color={
    draw=mapped color,
    fill=mapped color,
  },
  scatter,
  scatter src=explicit,
  mark options = fill] table [ x = hdlngth,y = skullw, z= Y_pred, meta expr= {(abs((\thisrow{Err_app_pred})))}] {possum_complete.txt};
 \end{axis}
\end{tikzpicture}}
\scalebox{0.9}{
\begin{tikzpicture}
\begin{axis}[
grid = both,
legend style={at={(0.5,.25)},anchor=north},
 legend style={draw=none},
ticklabel style = {font=\tiny, ultra thick},
y label style={ anchor=center },
xlabel=Head length,
y label style={rotate=75},
label style={font=\tiny},
zticklabel pos=left,
ylabel=Skull length,
zlabel = {Possum height},
grid style={ultra thin, gray!15},
x label style= {rotate=90},
  every axis x label/.append style={sloped, at={(rel axis cs: 0.5, -0.250, 0)}, below},
  every axis y label/.append style={sloped, at={(rel axis cs: 1.250, 0.5, 0)}, below}, 
view={45}{30},
height=6.85cm,width=6.85cm,
mark size=.7pt,
]
\addplot3+[only marks, mark size=0.7pt, mark = *,
  color = lightgray!30,
  mark options={fill=lightgray!30}
  ]
  table [ x index = 0,y index = 1, z index = 8  ] {possum_complete.txt};
  \addplot3+[only marks, colormap/YlOrRd,mark size=0.7pt, mark = *,scatter,scatter/use mapped
color={
    draw=mapped color,
    fill=mapped color,
  },
  scatter,
  point meta min = 0,
  scatter src=explicit,
        visualization depends on=\thisrowno{7}\as\wtwo,
    scatter/@pre marker code/.append style={
        /tikz/mark size=0.7*ln(exp(1)+50*abs(\wtwo))},
  mark options = fill]
    table [x index = 0, y index = 1, z index = 6,meta expr= {(abs(\thisrowno{7}))}]  {test_possum_expl.txt};
        \legend{data, selected},
\end{axis}
\end{tikzpicture}} \\
\scalebox{0.9}{
\begin{tikzpicture}
\begin{axis}[
grid = both,
ticklabel style = {font=\tiny, ultra thick},
colorbar horizontal,
colorbar style={
at={(0.45,0.25)},anchor=north },
   colorbar style={
         point meta min=0,  
         point meta max = 0.02,
         scaled x ticks = false,
        xtick = {0,  0.02},
            width = 20mm,
            title = Error,
            title style={xshift=-0.6cm,yshift=-0.25cm},
            },
         colorbar/width=2.5mm,
y label style={ anchor=center },
xlabel=Foot length,
y label style={rotate=75},
label style={font=\tiny},
zticklabel pos=left,
ylabel=Ear conch length,
zlabel = {Possum height},
grid style={ultra thin, gray!15},
x label style= {rotate=90},
  every axis x label/.append style={sloped, at={(rel axis cs: 0.5, -0.250, 0)}, below},
  every axis y label/.append style={sloped, at={(rel axis cs: 1.250, 0.5, 0)}, below}, 
colormap/YlOrRd,
point meta min = 0,
view={45}{30},
height=6.85cm,width=6.85cm,
mark size=.7pt,
]
\addplot3+[only marks, mark size=0.7pt, mark = *,scatter,scatter/use mapped
color={
    draw=mapped color,
    fill=mapped color,
  },
  scatter,
  scatter src=explicit,	
  mark options = fill] table [ x = footlgth,y = earconch, z= Y_pred, meta expr= {(abs(1e16*(\thisrow{Err_app_pred})))}] {possum_complete.txt};
 \end{axis}
\end{tikzpicture}}
\scalebox{0.9}{
\begin{tikzpicture}
\begin{axis}[
grid = both,
legend style={at={(0.5,.25)},anchor=north},
 legend style={draw=none},
ticklabel style = {font=\tiny, ultra thick},
y label style={ anchor=center },
xlabel=Foot length,
ylabel=Ear conch length,
y label style={rotate=75},
label style={font=\tiny},
zticklabel pos=left,
zlabel = {Possum height},
grid style={ultra thin, gray!15},
x label style= {rotate=90},
  every axis x label/.append style={sloped, at={(rel axis cs: 0.5, -0.250, 0)}, below},
  every axis y label/.append style={sloped, at={(rel axis cs: 1.250, 0.5, 0)}, below}, 
view={45}{30},
height=6.85cm,width=6.85cm,
mark size=.7pt,
]
\addplot3+[only marks, mark size=0.7pt, mark = *,
  color = lightgray!30,
  mark options={fill=lightgray!30}
  ]
  table [ x index = 2,y index = 3, z index = 8  ] {possum_complete.txt};
  \addplot3+[only marks, colormap/YlOrRd,mark size=0.7pt, mark = *,scatter,scatter/use mapped
color={
    draw=mapped color,
    fill=mapped color,
  },
  scatter,
          visualization depends on=\thisrowno{7}\as\wtwo,
    scatter/@pre marker code/.append style={
        /tikz/mark size=0.7*ln(exp(1)+50*abs(\wtwo))},
  point meta min = 0,
  scatter src=explicit,
  mark options = fill]
    table [x index = 2, y index = 3, z index = 6,meta expr= {(abs(\thisrowno{7}))}]  {test_possum_expl.txt};
    \legend{data, selected},
\end{axis}
\end{tikzpicture}} \\

\scalebox{0.9}{
\begin{tikzpicture}
\begin{axis}[
grid = both,
ticklabel style = {font=\tiny, ultra thick},
colorbar horizontal,
colorbar style={
at={(0.45,0.25)},anchor=north },
   colorbar style={
         point meta min=0,  
         point meta max =0.02,
         scaled x ticks = false,
        xtick = {0,  0.02},
            width = 20mm,
            title = Error,
            title style={xshift=-0.6cm,yshift=-0.25cm},
            },
         colorbar/width=2.5mm,
y label style={ anchor=center },
xlabel=Chest girth,
y label style={rotate=75},
label style={font=\tiny},
zticklabel pos=left,
ylabel=Belly girth,
zlabel = {Possum height},
grid style={ultra thin, gray!15},
x label style= {rotate=90},
  every axis x label/.append style={sloped, at={(rel axis cs: 0.5, -0.250, 0)}, below},
  every axis y label/.append style={sloped, at={(rel axis cs: 1.250, 0.5, 0)}, below}, 
colormap/YlOrRd,
point meta min = 0,
view={45}{30},
height=6.85cm,width=6.85cm,
mark size=.7pt,
]
\addplot3+[only marks, mark size=0.7pt, mark = *,scatter,scatter/use mapped
color={
    draw=mapped color,
    fill=mapped color,
  },
  scatter,
  scatter src=explicit,	
  mark options = fill] table [ x = chest ,y = belly, z= Y_pred, meta expr= {(abs((\thisrow{Err_app_pred})))}] {possum_complete.txt};
 \end{axis}
\end{tikzpicture}}
\scalebox{0.9}{
\begin{tikzpicture}
\begin{axis}[
grid = both,
legend style={at={(0.5,.25)},anchor=north},
 legend style={draw=none},
ticklabel style = {font=\tiny, ultra thick},
y label style={ anchor=center },
xlabel=Chest girth,
y label style={rotate=75},
label style={font=\tiny},
zticklabel pos=left,
ylabel=Belly girth,
zlabel = {Possum height},
grid style={ultra thin, gray!15},
x label style= {rotate=90},
  every axis x label/.append style={sloped, at={(rel axis cs: 0.5, -0.250, 0)}, below},
  every axis y label/.append style={sloped, at={(rel axis cs: 1.250, 0.5, 0)}, below}, 
view={45}{30},
height=6.85cm,width=6.85cm,
mark size=.7pt,
]
\addplot3+[only marks, mark size=0.7pt, mark = *,
  color = lightgray!30,
  mark options={fill=lightgray!30}
  ]
  table [ x index = 4,y index = 5, z index = 8  ] {possum_complete.txt};
  \addplot3+[only marks, colormap/YlOrRd,mark size=0.7pt, mark = *,scatter,scatter/use mapped
color={
    draw=mapped color,
    fill=mapped color,
  },
  scatter,
  point meta min = 0,
  scatter src=explicit,
          visualization depends on=\thisrowno{7}\as\wtwo,
    scatter/@pre marker code/.append style={
        /tikz/mark size=0.7*ln(exp(1)+50*abs(\wtwo))},
  mark options = fill]
    table [x index = 4, y index = 5, z index = 6,meta expr= {(abs(\thisrowno{7}))}]  {test_possum_expl.txt};
        \legend{data, selected},
\end{axis}
\end{tikzpicture}}
\caption{Possum dataset: scatter plots of the height of the animal versus pairs of correlated, standardized anatomical features. In the left plot, point colors are with respect to the observation-specific relative error: darker shades represent higher errors. In the right plot, selected data points are colored and sized based on the magnitude of their explanations: darker and bigger points indicate higher values.}
\label{fig:possum1} 
\end{center}
\end{figure}

Even though visually exploring multidimensional patterns can be challenging, Figure~\ref{fig:possum1} depicts the bivariate dependence of possum length with the corresponding head and skull structure, with the foot length and the size of the ear conch, and with the chest and belly dimensions. Notice that we have arranged the bidimensional relationships based on pairwise correlations, with the displayed pairs of explanatory variables being the most correlated ones. 
The results confirm that data points with the highest observation-specific explanations tend to be located in less dense areas and near the boundaries of the point cloud. It is also possible to infer that the most important possums for the prediction process are the numbers $24$ and $48$. The first animal is characterized by a narrow head (skull width = $54.9$), but longer feet (foot length = $75$) and ear conch (ear conch length = $53.5$). On the other hand, the second important possum has a wider head size (skull length = $59.6$), but shorter feet(foot length = $64$), and the smaller ear conch (ear conch length = $43.9$). A similar analysis can be conducted for all selected instances in the surrogate model. 
Therefore, we can conclude that observation-specific explanations highlight the most representative data instances (possums) for the reconstruction of the prediction process, discerning any redundancy of information.

\section{Discussion}\label{sec:discussion}
In this work, we shift the perspective of traditional explainability techniques from the importance of features to the importance of observations. We introduce the definition of observation-specific explanations, with the rationale of understanding the impact of each data point in the prediction process of a black-box model. Then, we propose a technique to obtain such observation-specific explanations. We employ scattered data approximation to define a surrogate model with provable high fidelity to the black box, using the orthogonal matching pursuit algorithm to estimate it. This procedure selects a subset of points to optimally define the surrogate model, which are then assigned individual coefficients. The observation-specific explanations are then obtained from these coefficients. Note that the proposed technique is completely model-agnostic: it can be applied to any black box of interest. 
Future research endeavors could explore the development of novel techniques for computing observation-specific explanations. This may involve alternative approaches to defining surrogate models or bypassing their definition altogether. Furthermore, there is potential for investigating the applicability of our approach to different and unstructured data types, such as images or texts, \textcolor{black}{or in different fields (finance, healthcare and so forth)}. Additionally, 
 studying the efficiency of the method in scenarios with high dimensionality or a large number of data points could provide valuable insights.

\newpage

%
%
\bibliographystyle{splncs04.bst}
\bibliography{bibliography}
\end{document}